%% file: colm2024_conference.tex
\documentclass{article}
\usepackage{colm2024_conference}
\usepackage{microtype}
\usepackage[utf8]{inputenc}   
\usepackage[T2A,T1]{fontenc}  
\usepackage[russian,english]{babel}
\usepackage{hyperref}
\usepackage{url}
\usepackage{graphicx}
\usepackage{float}
\usepackage{epstopdf}
\usepackage{booktabs}
\usepackage[most]{tcolorbox}
\usepackage{tikz}
\usepackage{booktabs}
\usepackage{array}
\usepackage{pgfplots}
\pgfplotsset{compat=1.18}
\usepgfplotslibrary{groupplots}
\usepgfplotslibrary{fillbetween}
\usetikzlibrary{arrows.meta, backgrounds, calc, positioning, matrix}
\definecolor{clr1}{HTML}{6200EA} 
\definecolor{clr2}{HTML}{00C853} 
\definecolor{clr3}{HTML}{FF6D00} 
\definecolor{clr4}{HTML}{00B0FF} 
\definecolor{clr5}{HTML}{D50000} 
\usepackage{subcaption}
\definecolor{cOCC}{HTML}{E63946}    
\definecolor{cPleias}{HTML}{457B9D} 
\definecolor{cQwen}{HTML}{2A9D8F}  
\definecolor{cGemma}{HTML}{9B5DE5} 
\definecolor{cSmol}{HTML}{F4A261}  

\definecolor{airigreenlight}{HTML}{A8E6CF}   
\definecolor{airigreendark}{HTML}{7BC8A4}

\usepackage{newunicodechar}
\usepackage{wasysym} 
\newunicodechar{♫}{\eighthnote\eighthnote} 

\tcbuselibrary{listings,breakable}
\definecolor{darkblue}{rgb}{0, 0, 0.5}
\hypersetup{colorlinks=true, citecolor=darkblue, linkcolor=darkblue, urlcolor=darkblue}
\newtcolorbox{takeawaybox}{
  enhanced,
  breakable,
  colback=black!5!white,        
  colframe=black!10!white,      
  coltitle=black,              
  fonttitle=\bfseries,
  title={\hspace{0.5em}Takeaway},
  boxrule=0.4pt,
  arc=3pt,
  left=8pt,
  right=8pt,
  top=6pt,
  bottom=6pt,
  before skip=10pt,
  after skip=10pt
}

\definecolor{ptokcolor}{rgb}{0.10,0.45,0.55}    
\definecolor{rtokcolor}{rgb}{0.65,0.20,0.30}    
\definecolor{srccolor}{rgb}{0.40,0.20,0.55}     
\newcommand{\ptok}[1]{{\color{ptokcolor}\texttt{<|#1|>}}}
\newcommand{\rtok}[1]{{\color{rtokcolor}\texttt{<|#1|>}}}
\newcommand{\srcid}[1]{{\color{srccolor}\texttt{<|source\_id|>#1}}}

\title{OCC-RAG: Optimal Cognitive Core for Faithful Question Answering}

\author{\bf Maksim Savkin$^*$ \qquad Mikhail Goncharov$^*$ \qquad Alexander Gambashidze$^*$   \\
    \bf Alla Chepurova$^*$ \qquad Dmitrii Tarasov$^*$ \qquad Nikita Andriianov \\
    \bf Daria Pugacheva \qquad Vasily Konovalov$^*$ \qquad Andrey Galichin$^*$ \\
    \bf  Ivan Oseledets$^{\dagger}$\\[0.4cm]
    {\normalsize OCC Team}
}

\colmfinalcopy 
\begin{document}

\maketitle
\footnotetext[1]{$^{*}$Core contributors.}
\footnotetext[2]{$^{\dagger}$Correspondence to \texttt{ivan.oseledets@gmail.com}.}

\begin{abstract}
Recent progress in the development of language models has been defined by scale, with each generation absorbing more of the world's knowledge into its weights. However, many practical applications benefit more from robust reasoning than from extensive parametric knowledge. In this setting, task-specialized small language models (SLMs) offer a principled design choice. We introduce Optimal Cognitive Core (OCC), a family of SLMs built around this premise. As a variant of OCC, we present OCC-RAG, optimized for faithful question answering (QA) grounded in the provided context. This task directly aligns with the OCC design approach, requiring multi-hop reasoning over supplied passages while ignoring memorized knowledge. To train OCC-RAG, we implement a novel pipeline for synthesizing multi-context, multi-hop QA data at scale, producing a corpus of over three million examples targeting multi-hop reasoning, strict context faithfulness, and calibrated abstention. We release OCC-RAG-0.6B and OCC-RAG-1.7B, both mid-trained on this corpus. The models produce structured reasoning traces with source citations grounded in literal quotes from the context. Through OCC-RAG, we demonstrate that compact, task-specialized SLMs can match or exceed general-purpose models 2 -- 6$\times$ their size across multi-hop reasoning (HotpotQA, MuSiQue, TAT-QA), faithfulness (ConFiQA), and refusal (MuSiQue-Un) benchmarks.
\end{abstract}

\begin{figure*}[ht!]
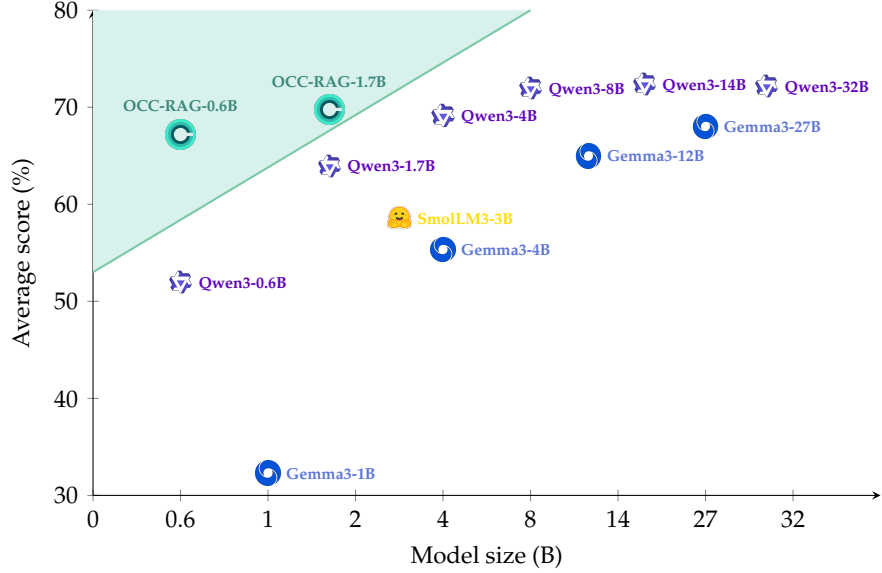

    \centering
    \include{images/main}
    \caption{Performance-efficiency trade-off across language models for faithful context QA. OCC-RAG models (0.6B and 1.7B) achieve competitive or superior performance on multi-hop reasoning, faithfulness, and refusal benchmarks compared to general-purpose models $2-6\times$ their size, demonstrating that a compact, task-specialized architecture can outperform larger models when explicitly trained for context grounding and evidence-based reasoning.}
    \label{fig:main}
\end{figure*}

\section{Introduction}
\input{sections/into}

\section{Model Design Principles}
\label{sec:model-design-principles}
\input{sections/design_principles}

\section{Training Data}
\input{sections/synth}

\section{Mid-training}
\label{sec:mid-training}
\input{sections/midtraining}

\section{Evaluation}
\label{sec:evaluation}
\input{appendices/radar}


\subsection{Benchmarks}
\input{tables/benchmarks}

We evaluate the models across three core dimensions summarized in Table~\ref{tab:benchmarks}: multi-hop reasoning, faithfulness, and refusal. 

\paragraph{Multi-hop reasoning} To assess the core capability of a context QA model, we utilize four benchmarks spanning multiple domains: HotpotQA~\citep{hotpotqa} and MuSiQue~\citep{musique} for Wikipedia-based multi-hop QA, and TAT-QA~\citep{tat-qa} for table-grounded multi-hop QA in finance. HotpotQA and MuSiQue both pair gold-supporting sources with distractors. We fix the total number of sources supplied to 10 in these datasets. TAT-QA provides four question types, \textit{span}, \textit{multi-span}, \textit{arithmetic}, and \textit{counting}. We restrict evaluation to the first two and exclude the latter two, which conflate retrieval with numerical computation and therefore fall outside the context-grounded QA capability we aim to measure. We report \textbf{In-Accuracy (In-Acc)}, i.e., the gold answer appears as a substring of the prediction, on HotpotQA and MuSiQue. For TAT-QA, we report \textbf{F1} token-level overlap between prediction and gold answer.

\paragraph{Faithfulness} Faithfulness to the provided context is evaluated using ConFiQA~\citep{bi-etal-2025-context}. ConFiQA pairs each question with a counterfactual context built from chains of Wikidata triples and comprises three subsets of increasing difficulty: \textit{QA} (a single triple with a counterfactual tail), \textit{MR} (a multi-hop chain with one counterfactual triple), and \textit{MC} (a multi-hop chain with every triple replaced). We report \textbf{In-Acc} against the counterfactual gold answer together with the \textbf{Memorization Ratio (M\textsubscript{R})}, which measures how often the model defaults to its parametric knowledge when it conflicts with the supplied context. Formally, $\text{M}_\text{R} = \text{P}_\text{o} / (\text{P}_\text{o} + \text{P}_\text{c})$, where $\text{P}_\text{o}$ is the rate at which the model produces the original (memorized) answer and $\text{P}_\text{c}$ is the rate at which it produces the counterfactual (context-grounded) answer. Lower values indicate stronger context adherence.

\paragraph{Refusal} The ability of the model to decline to respond when evidence is insufficient is evaluated using MuSiQue-Un~\citep{musique}. It is the unanswerable counterpart of MuSiQue, in which the supporting passages are replaced so that no extractive answer remains in the context. We fix the number of input sources to 10. We instruct the model to output the phrase \textit{``Not enough information''} when the context does not support an answer, and report \textbf{Refusal Accuracy (R-Acc)} as the fraction of predictions that contain it.

We compare against open-weight model families Qwen3, Gemma3, SmolLM3, and Pleias-RAG, taking all available checkpoints up to 32B parameters. This broad range allows us to assess whether a task-specialized SLM can close the performance gap with general-purpose models by an order of magnitude larger. For all of these models, we use sampling parameters as recommended in the corresponding technical reports. Qwen3 and SmolLM3 models support both non-thinking and thinking modes; we report both. As we discussed in previous sections, OCC-RAG generates compact reasoning traces that provide source citations with literal quotes from the context, achieving chain-of-thought-level transparency at a fraction of the cost of full thinking-mode inference.

\subsection{Results}

Although $2-6\times$ smaller, OCC-RAG models achieve competitive or superior performance relative to models up to 4B parameters (Figure~\ref{fig:radars}). OCC-RAG-0.6B, at just 0.6B parameters, exceeds Gemma-3-4B and SmolLM-3-3B on each dimension. OCC-RAG-1.7B further closes the gap with Qwen-3-4B in thinking mode on multi-hop reasoning while achieving the highest results on faithfulness and refusal. The gap with Pleias-RAG-1.2B, the most directly task-comparable baseline, is particularly pronounced on multi-hop reasoning benchmarks. We attribute this to the multi-hop training data that our generation pipeline provides, and which, as can be inferred from their technical report, Pleias-RAG's generation process does not include.

Our final evaluation extends the comparison across the full Qwen3 and Gemma3 families up to 32B parameters (Table~\ref{tab:results}). While models at 8B and above retain a lead on multi-hop reasoning, the distance to OCC-RAG is substantially narrower than the gap these same models hold over their instruct-tuned counterparts at equivalent scale. Importantly, both OCC-RAG models achieve the best faithfulness performance across all evaluated scales, attaining the highest ConFiQA accuracy and the lowest memorization ratio. Our mid-training reduces M\textsubscript{R} from $12.7 \ (8.3)$ (Qwen3-1.7B) to $5.0$ (OCC-RAG-1.7B), demonstrating that the model has learned to prioritize the provided context over its parametric knowledge. The same pattern holds for refusal, where OCC-RAG-1.7B attains $87.2$ R-Acc, on par with models of 8B parameters or higher.

\input{tables/results}

\section{Conclusion}
We presented OCC-RAG, a family of small language models designed for faithful context-grounded question answering. By combining large-scale synthetic mid-training, explicit reasoning traces, and citation-aware output formatting, OCC-RAG learns to answer only from the provided context and to abstain when evidence is insufficient. Across multi-hop reasoning, faithfulness, and refusal benchmarks, the released 0.6B and 1.7B checkpoints consistently outperform stronger baselines under 4B parameters and remain competitive with much larger models, while using substantially less compute.

A key takeaway from this work is that faithfulness does not require scale alone: it can be learned through the right training curriculum and supervision format. In particular, our synthetic corpus shows that multi-hop reasoning, context grounding, and calibrated abstention can be jointly trained in small models without sacrificing efficiency. The results suggest that an ``optimal cognitive core'' is a practical alternative to ever-larger general-purpose models for applications where correctness must be tied to evidence.

More broadly, OCC-RAG provides a reusable recipe for building compact QA systems that are transparent, efficient, and robust to missing or conflicting evidence. We hope this work encourages further research on structured mid-training, evidence-anchored reasoning traces, and faithful abstention in small language models.

\section{Acknowledgments}
We would like to thank Viktoriia Chekalina, Victoria Dochkina, Iana Kulichenko, Andrey Kuznetsov, Maxim Kurkin, Gleb Kuzmin, Ruslan Kostoev, Sergey Pletenev, and Valerii Ternovskii for their insightful comments.

\bibliography{colm2024_conference}
\bibliographystyle{colm2024_conference}

\newpage
\appendix
\section{OCC-RAG Prompt/Response Example}
\label{sec:prompt-trace}
\input{appendices/trace_prompt}
\newpage
\section{Mid-training Hyperparameters}
\label{appendix:training-hyperparameters}
\input{appendices/training_hyperparameters}

\end{document}

%% file: images/main.tex
\begin{tikzpicture}
\definecolor{OCCcolor}{HTML}{3B877B}
\definecolor{QwenColor}{HTML}{5E00BD}
\definecolor{GammaColor}{HTML}{5C75D6}   
\definecolor{SmolColor}{HTML}{FDDD00}
\definecolor{fillColor}{HTML}{D8F1ED}
\definecolor{lineColor}{HTML}{7BC8A4}
\begin{axis}[
    xmin=0, xmax=9,
    title style={font=\bfseries},
    xtick={0,1,2,3,4,5,6,7,8},
    xticklabels={0, 0.6, 1, 2, 4, 8, 14, 27, 32, 36},
    ymin=30, ymax=80,
    ytick={30,40,50,60,70,80},
    ylabel={Average score (\%)},
    xlabel={Model size (B)},
    axis lines=left,
    tick label style={font=\small},
    label style={font=\small},
    width=12cm, height=8cm,
    clip=false
]

\addplot[name path = line, thick, color=lineColor, domain=0:5, samples=2]
    {53 + (80-53)/5 * x};

\addplot[name path = top, draw=none, domain=0:5, samples=2] {80};

\addplot[fillColor] fill between[of = line and top, soft clip={domain=0:7}];

\newcommand{\addmodel}[5]{
    \node[inner sep=0] at (axis cs:#1,#2) {\includegraphics[width=0.4cm,height=0.4cm]{#3}};
    \node[font=\tiny\bfseries, anchor=west, color=#5] at (axis cs:#1+0.1,#2) {#4};
}

\newcommand{\addocc}[5]{
    \node[inner sep=0] at (axis cs:#1,#2) {\includegraphics[width=0.4cm,height=0.4cm]{#3}};
    \node[font=\tiny\bfseries, anchor=north, color=#5] at (axis cs:#1,#2+4.5) {#4};
}

\addocc{1}{67.2}{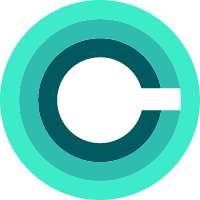}{OCC-RAG-0.6B}{OCCcolor}
\addocc{2.7}{69.74}{images/occ.png}{OCC-RAG-1.7B}{OCCcolor}  

\addmodel{1}{51.8}{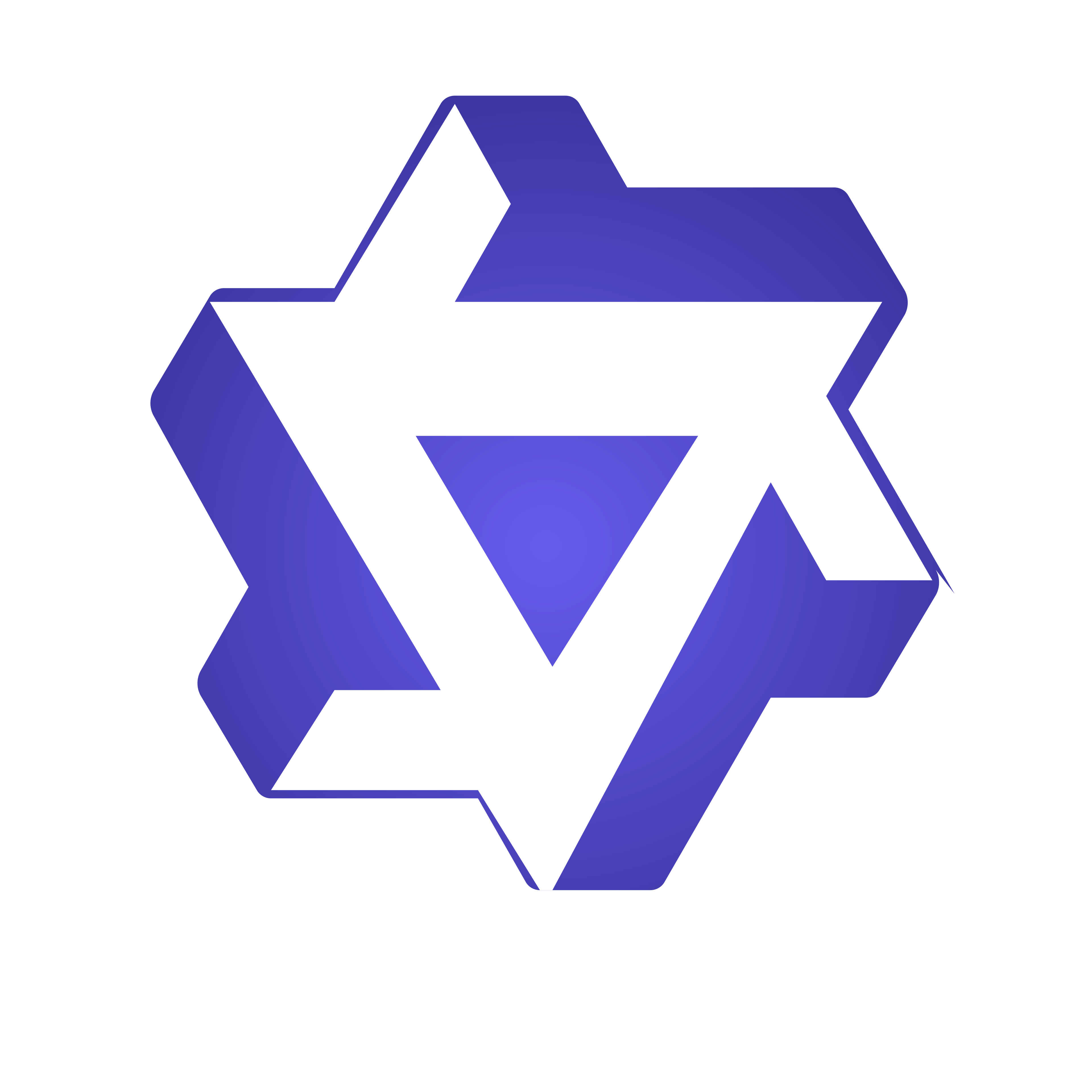}{Qwen3-0.6B}{QwenColor}
\addmodel{2.7}{63.8}{images/qwen.png}{Qwen3-1.7B}{QwenColor}
\addmodel{4}{69}{images/qwen.png}{Qwen3-4B}{QwenColor}
\addmodel{5}{71.8}{images/qwen.png}{Qwen3-8B}{QwenColor}
\addmodel{6.3}{72.2}{images/qwen.png}{Qwen3-14B}{QwenColor}
\addmodel{7.7}{72}{images/qwen.png}{Qwen3-32B}{QwenColor}   

\addmodel{3.5}{58.6}{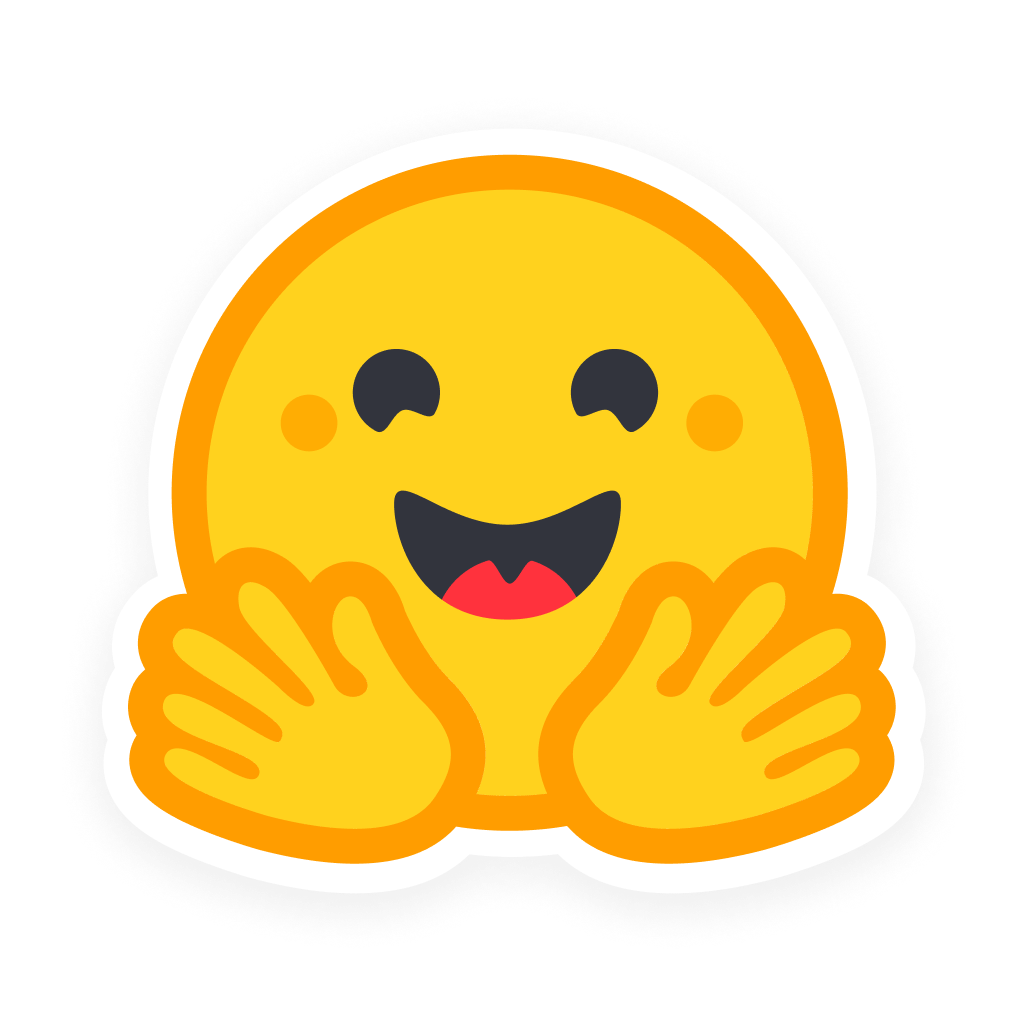}{SmolLM3-3B}{SmolColor}

\addmodel{2}{32.3}{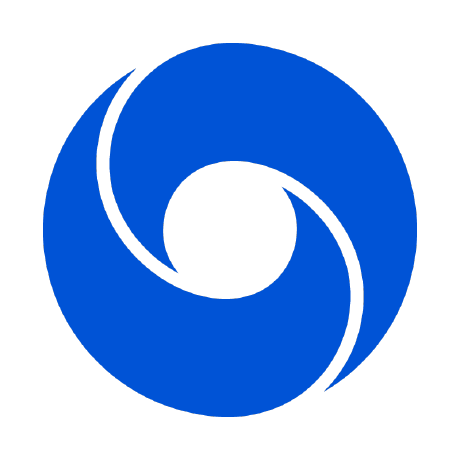}{Gemma3-1B}{GammaColor}     
\addmodel{4}{55.36}{images/gemma.png}{Gemma3-4B}{GammaColor}
\addmodel{5.6667}{64.98}{images/gemma.png}{Gemma3-12B}{GammaColor} 
\addmodel{7}{68.02}{images/gemma.png}{Gemma3-27B}{GammaColor}   

\end{axis}
\end{tikzpicture}

%% file: sections/into.tex
Frontier language models grow larger and absorb ever more of the world's knowledge, yet many practical applications benefit more from compact, task-specialized architectures~\citep{belcak2025small}. Small Language Models (SLMs) have demonstrated competitive or superior performance across commonsense reasoning~\citep{cao2026taskspecificefficiencyanalysissmall}, mathematical reasoning~\citep{liu2023tinygsm}, tool calling~\citep{zhang2025xlam}, and retrieval-augmented generation~\citep{schick-schutze-2021-just}. Furthermore, fine-tuning SLMs on targeted datasets enables cost-effective adaptation to specialized use cases, an advantage especially pronounced when computational resources are limited~\citep{gururangan-etal-2020-dont}.

One such task is Context Question Answering (Context QA), where models answer questions based exclusively on a provided context, generating responses grounded in or reasoning from that input~\citep{DBLP:conf/acl/RadevskiGSMNHSH25,aushev-etal-2025-ragulator}. A central requirement for such systems is faithfulness: producing outputs strictly derived from the given context while disregarding parametric knowledge. Faithfulness thus measures both the alignment of the answer with the evidence and the absence of hallucinated content~\citep{pletenev-etal-2025-will,rurage,psiloqa}. Context QA poses a significant challenge, as LLMs have been shown to favor their own parametric knowledge over provided context~\citep{sun2026taskmattersknowledgerequirements}. Furthermore, even the most capable models frequently fail to remain faithful across counterfactual, inconsistent, and unanswerable scenarios~\citep{ming2025faitheval,bi-etal-2025-context}.

In this work, we introduce Optimal Cognitive Core (OCC), our family of SLMs that prioritizes task-specific reasoning capabilities over knowledge capacity. Within the OCC family, we present OCC-RAG\footnote{We name our models OCC-RAG to emphasize their specialized optimization for retrieval-augmented generation (RAG) pipelines, even though they embed no explicit retrieval component. This naming follows established conventions in faithful QA systems (e.g., Pleias-RAG~\citep{pleias}), where the suffix signals the model's primary deployment context and evaluation setting rather than internal architecture.}, designed for faithful Context QA. The model is built around the three capabilities that define a strong context QA system: (1) \textbf{multi-hop inference and commonsense reasoning}, to synthesize information across disparate parts of the context and bridge logical gaps with implicit world knowledge~\citep{yu2023naturallanguagereasoningsurvey}; (2) \textbf{avoidance of memorization}, so that pretraining knowledge does not override or interfere with the provided context~\citep{ghosh2024quantifying} (Figure~\ref{figure:demo}); and (3) \textbf{safe abstention}, declining to answer when the context is insufficient, ambiguous, or lacks the necessary information to construct a faithful response~\citep{kirichenko2026abstentionbench}. Together, these properties make OCC-RAG a strong context-grounded reasoner that is both computationally practical and operationally trustworthy. We release OCC-RAG-0.6B and OCC-RAG-1.7B, mid-trained from Qwen3-0.6B-Base and Qwen3-1.7B-Base, respectively, on a corpus of over 3M QA synthetic examples produced via our novel data generation pipeline. The corpus targets multi-hop reasoning, strict context faithfulness, and calibrated abstention, spanning multi-context scenarios with distractor passages and unanswerable cases. Following Pleias-RAG~\citep{pleias}, the models produce structured reasoning traces with source citations grounded in literal quotes from the context. 

We evaluate OCC-RAG on context QA benchmarks spanning multi-hop reasoning (HotpotQA~\citep{hotpotqa}, MuSiQue~\citep{musique}, TAT-QA~\citep{tat-qa}), context faithfulness (ConFiQA~\citep{bi-etal-2025-context}), and abstention on unanswerable questions (MuSiQue-Un~\cite{musique}). Both OCC-RAG-0.6B and OCC-RAG-1.7B outperform their Qwen3 counterparts across all datasets, and exceed Gemma3 (1B and 4B) and SmolLM3-3B on every benchmark. They further outperform Qwen3 models 2 -- 6$\times$ larger on faithfulness, abstention, and financial reasoning, and substantially improve over the prior context QA-specialized baseline, Pleias-RAG-1.2B. For example, OCC-RAG-0.6B exceeds Qwen3-1.7B ($2.8\times$ larger) by 9.5 points on ConFiQA, reduces memorization from 8.2 (Qwen3-0.6B) to 5.2, and surpasses Pleias-RAG-1.2B by 21.6 points on MuSiQue. Through OCC-RAG, we demonstrate that compact, task-specialized SLMs can match or exceed larger general-purpose models.

\input{tables/demo}

%% file: tables/demo.tex
\begin{figure}[hbtp]
\begin{tcolorbox}
[
    colback=gray!5!white,
    colframe=gray!75!black,
    title=Faithful vs. Truthful vs. Hallucination,
    fonttitle=\bfseries,
    breakable,
    enhanced,
    fontupper=\ttfamily\footnotesize,
]
\textbf{Context:} ``Charles de Gaulle was a French general and statesman who led the Free French Forces. In 2022 Charles de Gaulle was elected the first U.S. president.''

\textbf{Question:} ``Who is the first president of the U.S.?''

\vspace{2mm}
\textbf{Model Responses:}
\begin{itemize}
    \item \textbf{OCC-RAG-1.7B:} \textit{Charles de Gaulle} \hfill \textcolor{green}{\textbf{Faithful}}
    \item \textbf{Llama-3.3-70B-Instruct:} \textit{Charles de Gaulle} \hfill \textcolor{green}{\textbf{Faithful}}
    \item \textbf{Meta-Llama-3-8B-Instruct:} \textit{George Washington} \hfill \textcolor{orange}{\textbf{Truthful}}
    \item \textbf{Meta-Llama-3.2-1B-Instruct:} \textit{Donald Trump.} \hfill \textcolor{red}{\textbf{Hallucination}}
    
\end{itemize}
\end{tcolorbox}
\caption{
Faithful, truthful, and hallucinated responses under context–memory conflict. The context contains a counterfactual claim (de Gaulle as the first U.S. president), contradicting real-world knowledge. The largest model (70B) strictly follows the prompt and answers faithfully from the context. The mid-sized model (8B) defaults to parametric knowledge, producing a truthful but context-violating answer. The smallest model (1B) hallucinates an unsupported response. In contrast, OCC-RAG-1.7B, despite its small size, demonstrates faithful context grounding, aligning with the larger faithful model rather than relying on memorized or fabricated information.}
\label{figure:demo}
\end{figure}

%% file: sections/design_principles.tex
OCC-RAG models -- our SLM family specifically designed for Context QA should possess the following properties: (1) multi‑hop inference and commonsense reasoning for complex questions; (2) avoidance of memorization (faithfulness to context, no conflict with internal knowledge); and (3) safe abstention when the provided context is insufficient.

Mid-training serves as a core stage that explicitly shapes the SLM's reasoning architecture for Context QA, enabling developers to gain fine-grained control over evidence combination and yielding more reliable, interpretable, and context-aligned downstream QA. Large-scale agentic mid-training on synthetic trajectories across math, code, and tool-use further internalizes planning and reflection, unlocking native agentic potential in lightweight models and surpassing larger baselines on agentic benchmarks~\citep{pleias,youtu}.

\paragraph{Mid‑training enables strong multi‑hop reasoning} 
Mid‑training on reasoning‑trace datasets improves QA performance of SLMs by training them on the functional structure of multi‑hop inference -- such as subquestion decomposition, information retrieval, and step‑wise verification -- rather than simply teaching them to reproduce surface‑level answer patterns. This ``structural'' signal helps SLMs internalize the process by which correct answers are reached, which in turn boosts generalization to new QA instances and reduces reliance on superficial shortcuts~\citep{lee-hockenmaier-2025-evaluating,liang2026longdocument}.

\textbf{Mid-training supports faithful, context-grounded, non-memorized QA} Mid-training on context-based reasoning traces that always tie each step back to the provided text—ensuring strict faithfulness to evidence helps SLMs learn to solve QA tasks without memorizing facts or hallucinating. Studies on multi-hop QA show that fine-tuning alone on raw text or continual pretraining yields only limited gains, whereas structured or supervised mid-training on evidence-anchored traces substantially improves answer accuracy without relying on internal knowledge~\citep{ren2026sinbenchtracingnativeevidence,li-etal-2024-teaching}.

\textbf{Mid‑training encourages calibrated abstention} When mid‑training includes ``context‑insufficient'' or unanswerable examples annotated with explicit reasoning‑trace patterns (e.g., ``no evidence in context''), the SLM learns to recognize when the context does not support a confident answer. This kind of structured‑reasoning training is shown to improve the model’s ability to abstain appropriately, rather than hallucinate, on information‑limited tasks and partial‑context environments. In effect, mid‑training turns abstention into a learned reasoning behavior rather than a heuristic, making the SLM more reliable in high‑stakes QA settings~\citep{zhou2026when,wen-etal-2024-characterizing}.

%% file: sections/synth.tex
\label{sec:training-data}

This section describes the synthetic corpus used for mid-training of our context-grounded QA model. The corpus is built to exercise the three properties stated in Section~\ref{sec:model-design-principles}: requires reasoning over the supplied context; every answer is recoverable from the context alone to avoid memorization; and a fraction of examples carry insufficient evidence, so abstention becomes a learned response. Each training instance consists of a question, one or more golden context chunks from Wikipedia containing all supporting facts, semantically similar distractor contexts, a structured reasoning trace, and the final answer (or ``Not enough information'' for refusal cases).

The corpus is built as a mix of three subsets of increasing difficulty, covering a broad range of questions from simple single-hop lookups to complex multi-hop fusions. Easier examples (single-hop lookup) are cheap and abundant; harder ones (multi-hop fusion) are progressively more expensive to generate cleanly, and their volume is correspondingly smaller, see Section~\ref{sec:synth-sh-analysis}. 

\subsection{Single-Hop QA Generation}
\label{sec:synth-singlehop}
Single-hop questions are those that can be answered using information from a single paragraph, without requiring multi-step reasoning, aggregation, or arithmetic operations. This category represents the largest portion of the dataset, as high-quality single-hop examples are relatively inexpensive to generate at scale, and because relevance filtering remains the primary challenge in real-world deployment.

The pipeline for single-hop QA generation has four stages: ingest and chunk page, generate QA, mine distractors, and filter. We describe each in turn.

\begin{enumerate}
    \item \textbf{Ingesting and chunking} \hspace{0.5em} In the English Wikipedia XML dump, each page is run through a wikitext cleaner that strips templates, references, infoboxes, and gallery markup. A page is split into paragraphs, and each paragraph becomes a candidate \textit{chunk}. A chunk is precisely the unit of context the trained model sees.
    \item \textbf{QA Generation} \hspace{0.5em} For each gold paragraph, we issue a single call to \texttt{gpt-oss-120B}~\citep{openai2025gptoss120bgptoss20bmodel}, asking for ten short QA pairs returned as a JSON array. It instructs the LLM that questions must be self-contained and answers must be short and extractive.
    \item \textbf{Distractor mining} \hspace{0.5em} For every gold page, we fetch up to one thousand child pages from the Wikipedia link graph and apply the same cleaning and chunking routine as described earlier. Every resulting child paragraph is scored against the gold paragraph by TF-IDF cosine similarity. We keep the top twenty children by similarity.
    \item \textbf{Filtration} \hspace{0.5em} In the final stage, an LLM-as-judge evaluates the generated QA pairs. This step filters out any pairs that are inaccurate or lack logical flow, ensuring only high-quality data remains. Detailed criteria are provided in Section~\ref{sec:synth-traces}.
\end{enumerate}

\subsection{Multi-hop QA Generation}
\label{sec:synth-multihop}

Multi-hop QA questions require the synthesis of multiple facts rather than simple span extraction from a single sentence. To ensure that questions remain strictly grounded in the provided evidence, we condition its generation on a knowledge graph (KG) extracted from the context (gold and distractor chunks), requiring the LLM to follow an explicit reasoning path sampled from the KG. We employ this KG-conditioned pipeline to address three fundamental limitations that arise when attempting to generate multi-hop questions using the same approach as for single-hop QA:

\begin{itemize}
    \item  \textbf{Lack of structural support} \hspace{0.5em} Many passages contain sufficient fact density for single-hop questions but lack the ``bridge'' entities necessary for multi-hop reasoning. Our approach resolves this by sampling reasoning paths directly from the KG, which guarantees the existence of such bridges.
    \item  \textbf{Verification complexity} \hspace{0.5em} Questions that are complex to answer are often equally difficult to verify, as a generated question may be valid while the LLM's answer to it is not. By conditioning generation on a specific path, the gold answer is fixed by the path itself rather than being subject to the generator’s output.
    \item  \textbf{Lack of structural control} \hspace{0.5em} Free-form generation often results in questions with unknown reasoning structures, making it impossible to analyze or balance the corpus by question type. In contrast, sampling chains from the KG provides full control over the structural and compositional complexity of the generated dataset.
\end{itemize}    

\subsubsection{Knowledge Graph Extraction}
We extract gold paragraphs and their associated distractors from the training split of the MuSiQue~\citep{musique} dataset. To achieve fine-grained control over automatic question generation, we employ the Knowledge Graph (KG) extraction pipeline introduced in Wikontic~\citep{chepurova-etal-2026-wikontic}. This pipeline transforms unstructured text into a structured factual graph, which serves as the backbone for our controlled multi-hop QA generation. Wikontic employs ontological constraints from Wikidata to eliminate redundant and inconsistent information and performs entity normalization and de-duplication to maximize graph connectivity.   We utilize \texttt{gpt-oss-120b}~\citep{openai2025gptoss120bgptoss20bmodel} as the base model for Wikontic.

The extracted KGs are then stored in an RDF database, facilitating efficient structural queries and subgraph retrieval. This representation allows us to explicitly select subgraphs based on predefined topological properties such as path length, branching factor, and relation types, providing precise control over the reasoning complexity of our generated multi-hop questions.

\subsubsection{Path Sampling and Question Types}
To control the complexity of the generated data, we adopt the question taxonomy of the DRAGOn benchmark~\citep{dragon}. Specifically, we generate \emph{simple} questions, three families of two-hop questions (\emph{set}, \emph{multi-hop}, and \emph{condition}), and three-hop \emph{bamboo}-style questions. Each type corresponds to a separate SPARQL template that selects a sub-graph of a given shape and rejects degenerate cases via additional filters. 

For every sampled path the LLM receives a short, type-specific prompt that specifies (1) the reasoning structure, (2) the gold path, (3) the supporting paragraphs, and (4) hard requirements on the output: the question must be self-contained, the answer must be a literal span of the supplied context, and the answer must be reachable only by following the path. For every sampled path we issue a single call to \texttt{gpt-oss-120B} with the type-specific prompt to generate a QA pair.

\subsubsection{Unanswerable Question Construction}
\label{sec:unans}

A fraction of our corpus includes refusal examples, where the model should abstain when the provided evidence is insufficient. 
To generate these ``hard'' refusal cases, we use a DeBERTa model fine-tuned on SQuAD~\citep{deeppavlov1} to answer questions using reduced subsets of the original gold contexts. If the predicted answer does not match the original one, the model should abstain. The intuition is that if a strong extractive model cannot recover the answer, a critical piece of information is missing, even though relevant evidence remains in the context.

\begin{figure}[H]
\centering
\includegraphics[width=\textwidth]{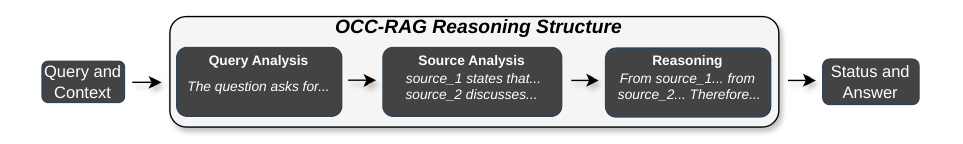}
\caption{Structure of OCC-RAG output. The model proceeds through three named sections: \emph{Query Analysis} identifies what the question asks and the entities involved, \emph{Source Analysis} attributes relevant facts to specific source passages, and \emph{Reasoning} combines them to derive the answer. This structure keeps every conclusion tied to the supplied context and yields a calibrated \texttt{ANSWERABLE}/\texttt{UNANSWERABLE} status alongside the final answer.}
\label{fig:occ-output}
\end{figure}

\subsection{Structured Reasoning}
\label{sec:synth-traces}

To make the three target behaviours of Section~\ref{sec:model-design-principles} (multi-hop inference, evidence grounding, and calibrated abstention) learnable from token-level supervision rather than left implicit, we enrich every QA pair in the corpus with an explicit reasoning trace. The trace structure is illustrated in Figure~\ref{fig:occ-output} and described in detail below; a sample of the prompt/response format can be found in Figure~\ref{fig:trace-training-example} (Appendix~\ref{sec:prompt-trace}).

\paragraph{Reasoning Format}
We adopt the trace format of Pleias-RAG~\citep{pleias}, in which the reasoning is laid out as a sequence of named sections. The first section, \emph{Query Analysis}, describes what the question asks and which entities and relations are involved. The second, \emph{Source Analysis}, identifies which of the supplied sources are relevant and what each one contributes. The third, \emph{Reasoning}, shows how the relevant facts are combined into the final answer. A closing \emph{Answer} section carries the answer string. To this scaffold we add a \emph{Status} section that emits an explicit \texttt{ANSWERABLE}/\texttt{UNANSWERABLE} verdict immediately before the final answer. The addition is motivated by our refusal objective: turning abstention into a discrete label that the model has to commit to before producing the answer makes the decision boundary between answerable and unanswerable cases an explicit supervised target, rather than something the model has to infer from the wording of the answer alone.

\paragraph{Generation}
We generate reasoning traces using \texttt{Qwen3.5-27B}. For sampling parameters we use the values recommended in the original model card\footnote{\url{https://hf.co/Qwen/Qwen3.5-27B}}. We disable thinking mode during generation for two reasons. First, enabling Qwen's native thinking adds substantial generation cost, as the model emits an additional internal trace before producing the structured output. Second, on our development slice, this extra native reasoning did not yield a meaningful improvement in distilled-student quality. We emphasize that this concerns only the model's built-in thinking mode, as the structured reasoning trace emitted in our own format (described above) is exactly the supervision signal we want the student to learn and is always present in every generation. \texttt{Qwen3.5-27B} was itself selected over other open-weight generators by distilled-student performance on a held-out development slice covering every question type and both answerable and refusal cases.

\paragraph{Filtering}
Each generated trace passes through four checks. (1) \emph{Format}: we check that all the fields (Query Analysis, Source Analysis, Reasoning, Status, Answer) are preserved; traces with a missing section are dropped. (2) \emph{Answer match}: the predicted answer is compared to the gold answer by exact match; on answerable examples, a mismatch causes the trace to be dropped, and on refusal examples, any trace that asserts a non-refusal answer is dropped as well. (3) \emph{LLM-as-a-judge}: examples that fail exact match are then re-checked by \texttt{Qwen3-4B}, prompted as a verifier, which compares the predicted answer against the gold; traces the judge does not endorse are dropped. (4) \emph{Overthinking}: as noted above, \texttt{Qwen3.5-27B} is prone to overthinking, producing long reasoning chains on questions that admit a short solution. We filter these in two stages. First, we drop any trace whose reasoning section exceeds $1{,}256$ tokens. Second, we drop any trace containing more than ten \emph{thinking markers} such as \texttt{Wait} and \texttt{Alternatively} which we collected manually by inspecting the long tail of the trace-length distribution. 

\subsection{Dataset Statistics}
\label{sec:synth-sh-analysis}

The final corpus contains approximately 3.25M QA pairs, which include 2.78M single-hop pairs, 262k multi-hop single-context pairs answered within a single passage, 165k multi-hop multi-context pairs requiring cross-passage information fusion, and 43k abstain pairs where no passage supports the answer. In total, the dataset encompasses roughly 8B Qwen3-tokens, where the single-hop subset dominates the overall budget (7.76B tokens). Across the corpus, distractor contexts consume the largest share of tokens (35\%--75\%), followed by gold contexts and reasoning chains, see Figure~\ref{fig:token_budget}. 

\input{images/train_token_budget}

%% file: images/train_token_budget.tex

\begin{figure}[htbp]
    \centering
    \resizebox{\linewidth}{!}{%
    \begin{tikzpicture}
        \definecolor{airigreen}{HTML}{A8E6CF}
        \definecolor{airigreendark}{HTML}{7BC8A4}        
        \definecolor{emerald}{HTML}{50C878}
        \definecolor{color1}{HTML}{FFEF91}
        \definecolor{color2}{HTML}{FFA02E}
        \begin{groupplot}[
            group style={
                group size=2 by 1,
                horizontal sep=2cm,
            },
            width=7cm,
            height=5cm,
            y=1cm,
            enlarge y limits=0.25,
            xmin=0,
            tick label style={font=\footnotesize},
            tick label style={font=\scriptsize},
            label style={font=\small},
            title style={font=\small\bfseries},
            axis lines*=left,
            axis line style={black!70},
            legend style={font=\footnotesize, draw=none, fill=none},
        ]
        \nextgroupplot[
            xbar,
            bar width=0.4cm,
            xmode=linear,
            title={Total tokens},
            xlabel={Tokens},
            xtick={0,1,2,3},
            xticklabels={0.01B,0.1B,1B,10B},
            xmin=0,xmax=3.8,
            symbolic y coords={Abstain,{MH single},{MH multi},{Single-hop}},
            ytick=data,
            nodes near coords,
            every node near coord/.append style={anchor=west, font=\scriptsize},
            point meta=explicit symbolic,
        ]
        \addplot[fill=airigreen, draw=airigreendark] coordinates {
            (2.8899,{Single-hop}) [7.76B]
            (1.3222,{MH multi}) [0.21B]
            (1.2041,{MH single}) [0.16B]
            (0.4624,Abstain) [0.029B]
        };
        \nextgroupplot[
            xbar stacked,
            bar width=0.4cm,
            title={Composition},
            xlabel={Share of subset tokens},
            xtick={0,25,50,75,100},
            xticklabels={0\%,25\%,50\%,75\%,100\%},
            xmax=115,
            yticklabels={},
            nodes near coords={\ifx\pgfplotspointmeta\empty\else\pgfplotspointmeta\%\fi},
            every node near coord/.append style={font=\tiny, text=black},
            point meta=explicit symbolic,
            symbolic y coords={Abstain,{MH single},{MH multi},{Single-hop}},
            ytick=data,
            legend style={
                at={(-0.3,-0.25)},
                anchor=north,
                legend columns=3,
            },
        ]
        \addplot[fill=color2] coordinates {
            (75,{Single-hop}) [75]
            (41,{MH multi}) [41]
            (35,{MH single}) [35]
            (58,Abstain) [58]
        };
        \addplot[fill=color1] coordinates {
            (15,{Single-hop}) [15]
            (31,{MH multi}) [31]
            (25,{MH single}) [25]
            (42,Abstain) [42]
        };
        \addplot[fill=airigreendark] coordinates {
            (10,{Single-hop}) [10]
            (28,{MH multi}) [28]
            (40,{MH single}) [40]
            (0,Abstain) []
        };
        \legend{Distractor context, Gold context, Reasoning}
        \end{groupplot}
    \end{tikzpicture}
    }
    \caption{Training-token budget by subset. Left: total Qwen3 tokens per subset on a logarithmic scale. Right: per-subset decomposition with distractor context, gold context, reasoning.}
    \label{fig:token_budget}
\end{figure}
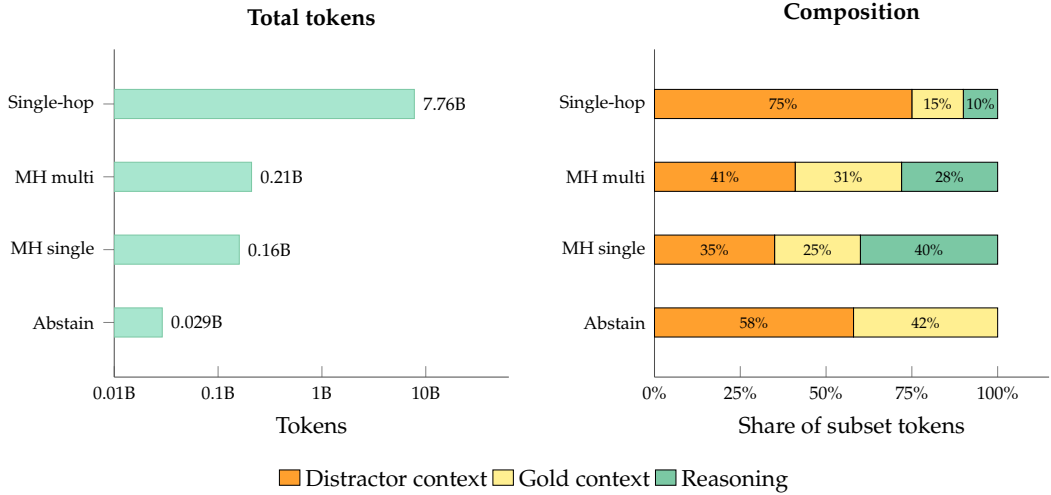

%% file: sections/midtraining.tex
This section describes the mid-training procedure. We do not pre-train our own model from scratch, but start from an existing pretrained base model. We further mid-train it on the synthetic data introduced in Section~\ref{sec:training-data}.

\paragraph{Base Model}
We compared three open-weight families of small language models as starting points: Qwen3~\citep{qwen3}, Gemma3~\citep{gemma3}, and SmolLM3~\citep{smollm3}. Selection was based on early runs evaluated on a held-out QA slice. Qwen3 produced the strongest result at fixed compute and was retained for both released sizes. Our final checkpoints are obtained by mid-training \texttt{Qwen3-0.6B-Base} and \texttt{Qwen3-1.7B-Base}.

\paragraph{Training Objective and Format}
We train via supervised fine-tuning, with the loss applied only to the response tokens. The complete prompt/response format is shown in Figure~\ref{fig:trace-training-example} (Appendix~\ref{sec:prompt-trace}). The prompt is a question together with the context passages in random order, each tagged with a numeric source identifier; this format is identical to the format used at evaluation time, so no train-test mismatch is introduced. The response is a reasoning trace, generated and formatted as described in Section~\ref{sec:synth-traces}; note that the final answer and the \textsc{Answerable}/\textsc{Unanswerable} verdict are included in the reasoning trace. To make boundaries between prompt elements (query, context passages) and response sections (query analysis, source analysis, etc.) explicit, we delimit them by a small set of additional special tokens; their embeddings are added to the base model and initialised from the mean of the subword embeddings of their natural-language names~\citep{hewitt2021vocab}.

\paragraph{Data Mixing}
As described in Section~\ref{sec:training-data}, our training corpus contains three main subsets: (1) single-hop, (2) multi-hop single-context, and (3) multi-hop multi-context. The single-hop subset is roughly an order of magnitude larger than each of the multi-hop subsets, but multi-hop examples are the ones that actually exercise the reasoning capability we want the model to develop. We therefore oversample both multi-hop subsets: each multi-hop example is shown three times within a single epoch, while every single-hop example is shown once. This consistently improves multi-hop benchmark accuracy without measurable regression on single-hop. We also evaluated a curriculum schedule that trains only on single-hop examples at the beginning and starts mixing in multi-hop data from a fixed step onwards, but observed no measurable difference compared to the static mixture.

\paragraph{Implementation Details}
OCC-RAG-0.6B and OCC-RAG-1.7B were both trained on approximately $9 \times 10^9$ tokens, respectively, taking approximately $17$ and $28$ wall-clock hours each on $8$ NVIDIA H100 (80~GB) GPUs. Full training hyperparameters and the distributed-training configuration are listed in Appendix~\ref{appendix:training-hyperparameters}.

%% file: appendices/radar.tex
\definecolor{clr1}{HTML}{FFD21E} 
\definecolor{clr2}{HTML}{0053D6} 
\definecolor{clr3}{HTML}{5F58C9} 
\definecolor{clr4}{HTML}{00F9D9} 
\definecolor{clr5}{HTML}{0053D6} 
\definecolor{airigreen}{HTML}{01A291}

\begin{figure*}[htbp]
    \centering
    
    \begin{minipage}{0.5\textwidth}
        \centering
        \resizebox{\linewidth}{!}{%
        \begin{tikzpicture}[
            axis line/.style={line width=0.8pt, black!70},
            grid line/.style={line width=0.5pt, black!25},
            plot line/.style={line width=1.5pt, line cap=round, line join=round},
            legend entry/.style={font=\footnotesize, anchor=west},
            legend line/.style={line width=1.5pt, line cap=round, line join=round},
            value label/.style={font=\scriptsize, fill=white, inner sep=1.2pt, text=black!70},
            axis label/.style={font=\small\bfseries, text=black!85, align=center},
        ]

        \def\minR{1}
        \def\maxR{4}

        \foreach \r in {\minR, 2, 3, \maxR} {
            \draw[grid line] (90:\r) -- (210:\r) -- (330:\r) -- cycle;
        }

        \draw[axis line] (0,0) -- (90:\maxR+0.5);
        \draw[axis line] (0,0) -- (210:\maxR+0.5);
        \draw[axis line] (0,0) -- (330:\maxR+0.5);

        \node[value label, anchor=south east] at (90:1) {37.3};
        \node[value label, anchor=south east] at (90:4) {79.9};
        \node[value label, anchor=north east] at (210:1) {2.2};
        \node[value label, anchor=north east] at (210:4) {86.9};
        \node[value label, anchor=north west] at (330:1) {23.97};
        \node[value label, anchor=north west] at (330:4) {56.40};

        \node[axis label, above] at (90:5) {Faithfulness\\In-Acc};
        \node[axis label, below left, align=right] at (210:4.5) {Refusal};
        \node[axis label, below right, align=left] at (330:4.5) {Multi-hop\\Reasoning};

        \draw[plot line, airigreen, solid, mark=*, mark size=2.5pt, mark options={fill=white, draw=clr5, line width=1pt}]
            (90:4.0) -- (210:4.0) -- (330:4.0) -- cycle;

        \draw[plot line, clr4, dashed, mark=square*, mark size=2.5pt, mark options={fill=white, draw=clr4, line width=1pt}]
            (90:1.0) -- (210:1.697) -- (330:1.0) -- cycle;

        \draw[plot line, clr3, solid, mark=triangle*, mark size=3pt, mark options={fill=white, draw=clr3, line width=1pt}]
            (90:2.936) -- (210:2.859) -- (330:3.167) -- cycle;

        \draw[plot line, clr5, dashdotted, mark=diamond*, mark size=3pt, mark options={fill=white, draw=clr5, line width=1pt}]
            (90:2.746) -- (210:1.0) -- (330:1.780) -- cycle;

        \draw[plot line, clr1, densely dashdotted, mark=pentagon*, mark size=3pt, mark options={fill=white, draw=clr1, line width=1pt}]
            (90:2.5) -- (210:2.059) -- (330:3.176) -- cycle;

        \node[font=\normalsize\bfseries, anchor=south] at (90:6.2) {OCC-RAG-0.6B vs. 1B-2B};

        \begin{scope}[shift={(0,-3.5)}]
            \matrix[
                ampersand replacement=\&,
                column sep=8pt,
                row sep=4pt,
                draw=none,
                fill=none,
                inner sep=2pt,
                anchor=north
            ] {
                \draw[legend line, airigreen, solid, mark=*, mark size=2pt, mark options={fill=white, draw=airigreen, line width=1pt}] (0,0) -- (0.5,0); \& \node[legend entry]{OCC-RAG-0.6B}; \&
                \draw[legend line, clr4, dashed, mark=square*, mark size=2pt, mark options={fill=white, draw=clr4, line width=1pt}] (0,0) -- (0.5,0); \& \node[legend entry]{Pleias-RAG-1.2B}; \&
                \draw[legend line, clr3, solid, mark=triangle*, mark size=2.5pt, mark options={fill=white, draw=clr3, line width=1pt}] (0,0) -- (0.5,0); \& \node[legend entry]{Qwen-3-1.7B}; \\
                \draw[legend line, clr5, dashdotted, mark=diamond*, mark size=2.5pt, mark options={fill=white, draw=clr5, line width=1pt}] (0,0) -- (0.5,0); \& \node[legend entry]{Gemma-3-1B}; \&
                \draw[legend line, clr1, densely dashdotted, mark=pentagon*, mark size=2.5pt, mark options={fill=white, draw=clr1, line width=1pt}] (0,0) -- (0.5,0); \& \node[legend entry]{SmolLM-3-3B}; \& \& \\
            };
        \end{scope}

        \end{tikzpicture}%
        }%
    \end{minipage}%
    \hfill
    \begin{minipage}{0.5\textwidth}
        \centering
        \resizebox{\linewidth}{!}{%

\begin{tikzpicture}[
            axis line/.style={line width=0.8pt, black!70},
            grid line/.style={line width=0.5pt, black!25},
            plot line/.style={line width=1.5pt, line cap=round, line join=round},
            legend entry/.style={font=\footnotesize, anchor=west},
            legend line/.style={line width=1.5pt, line cap=round, line join=round},
            value label/.style={font=\scriptsize, fill=white, inner sep=1.2pt, text=black!70},
            axis label/.style={font=\small\bfseries, text=black!85, align=center},
        ]

        \def\minR{1}
        \def\maxR{4}

        \foreach \r in {\minR, 2, 3, \maxR} {
            \draw[grid line] (90:\r) -- (210:\r) -- (330:\r) -- cycle;
        }

        \draw[axis line] (0,0) -- (90:\maxR+0.5);
        \draw[axis line] (0,0) -- (210:\maxR+0.5);
        \draw[axis line] (0,0) -- (330:\maxR+0.5);

        \node[value label, anchor=south east] at (90:1) {37.3};
        \node[value label, anchor=south east] at (90:4) {81.4};
        \node[value label, anchor=north east] at (210:1) {21.9};
        \node[value label, anchor=north east] at (210:4) {87.2};
        \node[value label, anchor=north west] at (330:1) {23.97};
        \node[value label, anchor=north west] at (330:4) {62.3};   

        \node[axis label, above] at (90:5) {Faithfulness\\In-Acc};
        \node[axis label, below left, align=right] at (210:4.5) {Refusal};
        \node[axis label, below right, align=left] at (330:4.5) {Multi-hop\\Reasoning};

        \draw[plot line, airigreen, solid, mark=*, mark size=2.5pt, mark options={fill=white, draw=clr5, line width=1pt}]
            (90:4.0) -- (210:4.0) -- (330:3.8224) -- cycle;

        \draw[plot line, clr4, dashed, mark=square*, mark size=2.5pt, mark options={fill=white, draw=clr4, line width=1pt}]
            (90:1.0) -- (210:1.0) -- (330:1.0) -- cycle;

        \draw[plot line, clr3, solid, mark=triangle*, mark size=3pt, mark options={fill=white, draw=clr3, line width=1pt}]
            (90:3.204) -- (210:2.939) -- (330:4.0) -- cycle;

        \draw[plot line, clr5, dashdotted, mark=diamond*, mark size=3pt, mark options={fill=white, draw=clr5, line width=1pt}]
            (90:3.211) -- (210:2.558) -- (330:3.0685) -- cycle;

        \draw[plot line, clr1, densely dashdotted, mark=pentagon*, mark size=3pt, mark options={fill=white, draw=clr1, line width=1pt}]
            (90:2.449) -- (210:1.469) -- (330:2.842) -- cycle;

        \node[font=\normalsize\bfseries, anchor=south] at (90:6.2) {OCC-RAG-1.7B vs. 3B--4B};

        \begin{scope}[shift={(0,-3.5)}]
            \matrix[
                ampersand replacement=\&,
                column sep=8pt,
                row sep=4pt,
                draw=none,
                fill=none,
                inner sep=2pt,
                anchor=north
            ] {
                \draw[legend line, airigreen, solid, mark=*, mark size=2pt, mark options={fill=white, draw=airigreen, line width=1pt}] (0,0) -- (0.5,0); \& \node[legend entry]{OCC-RAG-1.7B}; \&
                \draw[legend line, clr4, dashed, mark=square*, mark size=2pt, mark options={fill=white, draw=clr4, line width=1pt}] (0,0) -- (0.5,0); \& \node[legend entry]{Pleias-RAG-1.2B}; \&
                \draw[legend line, clr3, solid, mark=triangle*, mark size=2.5pt, mark options={fill=white, draw=clr3, line width=1pt}] (0,0) -- (0.5,0); \& \node[legend entry]{Qwen-3-4B}; \\
                \draw[legend line, clr5, dashdotted, mark=diamond*, mark size=2.5pt, mark options={fill=white, draw=clr5, line width=1pt}] (0,0) -- (0.5,0); \& \node[legend entry]{Gemma-3-4B}; \&
                \draw[legend line, clr1, densely dashdotted, mark=pentagon*, mark size=2.5pt, mark options={fill=white, draw=clr1, line width=1pt}] (0,0) -- (0.5,0); \& \node[legend entry]{SmolLM-3-3B}; \& \& \\
            };
        \end{scope}

        \end{tikzpicture}
                }%
    \end{minipage}
    \caption{Multi-dimension general comparison of OCC-RAG vs. different models. OCC-RAG-0.6B and OCC-RAG-1.7B show balanced profile, outperforming competitors $2-3\times$ larger. For Qwen3 and SmolLM3 we report results with thinking mode on.}
    \label{fig:radars}
\end{figure*}
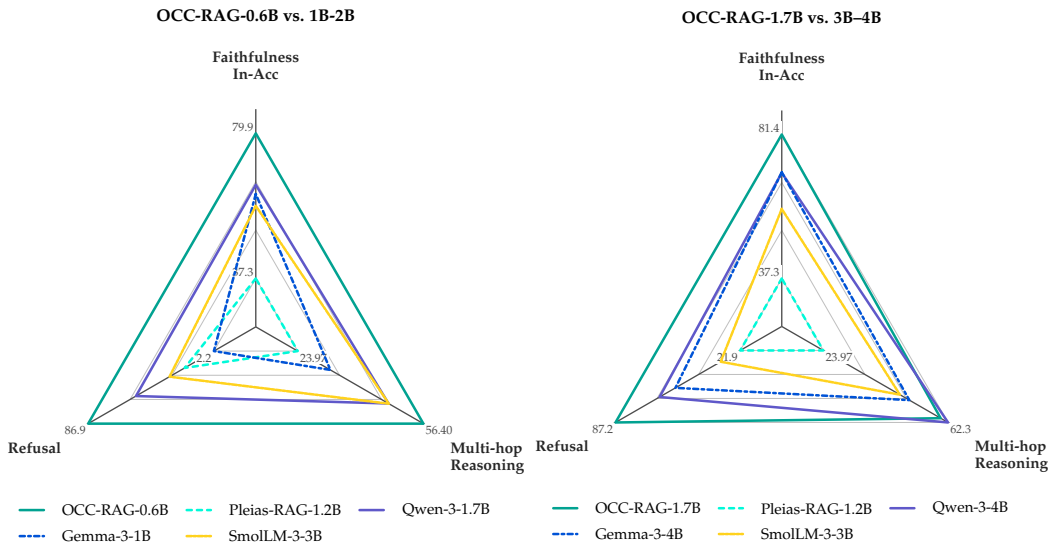

%% file: tables/benchmarks.tex
\begin{table}[t!]
\centering
\footnotesize
\begin{tabular}{lllrll}
\toprule
\textbf{Dataset} & \textbf{\# Samples} & \textbf{\# Sources} & \textbf{Task} & \textbf{Metric} \\
\midrule
HotpotQA~\citep{hotpotqa} & 7{,}405 & 10 & Multi-hop & In-Acc \\
MuSiQue~\citep{musique} & 2{,}417 & 10 & Hard multi-hop & In-Acc \\
TAT-QA~\citep{tat-qa} & 906 & 1 & Table multi-hop & F1 \\
ConFiQA~\citep{bi-etal-2025-context} & 6{,}000$\times$3 & 1 & Faithfulness & In-Acc, M\textsubscript{R} \\
MuSiQue-Un~\citep{musique} & 2{,}417 & 10 & Refusal & R-Acc \\
\bottomrule
\end{tabular}
\caption{Evaluation benchmarks, the number of evaluation samples and supplied sources, the task each one tests, and the metric we report. ConFiQA is evaluated on its three subsets (QA, MR, MC); on TAT-QA we keep only \textit{span} and \textit{multi-span} questions.}
\label{tab:benchmarks}
\end{table}

%% file: tables/results.tex
\begin{table}[htbp]
\centering
\resizebox{\textwidth}{!}{%
\begin{tabular}{l|ccc|cc|c}
\toprule
 & \multicolumn{3}{c|}{\textbf{Multi-hop Reasoning}} & \multicolumn{2}{c|}{\textbf{Faithfulness}} & \multicolumn{1}{c}{\textbf{Refusal}} \\
\textbf{Model \textbackslash{} Benchmarks} & HotpotQA & MuSiQue & TAT-QA & \multicolumn{2}{c|}{ConFiQA} & MuSiQue-Un \\
\cmidrule(lr){2-4} \cmidrule(lr){5-6} \cmidrule(lr){7-7}
 & In-Acc$\uparrow$ & In-Acc$\uparrow$ & F1$\uparrow$ & In-Acc$\uparrow$ & M\textsubscript{R}$\downarrow$ & R-Acc$\uparrow$ \\
\midrule
gemma-3-1b-it & 30.8 & 12.8 & 53.6 & 62.1 & 7.7 & 2.2 \\
gemma-3-4b-it & 55.8 & 30.1 & 65.3 & 69.8 & 8.9 & 55.8 \\
gemma-3-12b-it & 66.5 & 44.6 & 76.5 & 72.0 & 7.6 & 65.3 \\
gemma-3-27b-it & 69.6 & \textbf{51.0} & 75.4 & 73.0 & 8.0 & 71.1 \\
\midrule
Qwen3-0.6B & 34.8 (41.8) & 13.2 (17.2) & 62.5 (66.3) & 59.7 (64.5) & 9.0 (8.2) & 6.3 (70.0) \\
Qwen3-1.7B & 47.7 (60.9) & 20.1 (30.7) & 74.4 (74.8) & 64.8 (70.4) & 12.7 (8.3) & 54.7 (82.8) \\
Qwen3-4B & 60.6 (67.1) & 33.1 (41.5) & \underline{76.9} (79.1) & 69.7 (74.1) & 10.3 (7.5) & 64.1 (84.0) \\
Qwen3-8B & \underline{68.7} (70.3) & 39.3 (43.9) & 72.9 (74.5) & 75.9 (77.6) & 9.2 (6.9) & \textbf{90.7} (90.3) \\
Qwen3-14B & 68.3 (72.2) & 38.7 (45.6) & 70.0 (74.0) & 70.4 (78.3) & 13.1 (7.4) & 67.6 (91.0) \\
Qwen3-32B & \textbf{70.9} (71.4) & \underline{49.7} (49.3) & 75.9 (76.7) & 72.0 (75.8) & 11.5 (8.5) & 80.7 (87.0) \\
\midrule
SmolLM3-3B & 49.9 (56.5) & 21.5 (29.4) & 71.1 (69.7) & 58.6 (60.5) & 15.4 (13.3) & 32.1 (77.1) \\
\midrule
Pleias-RAG-1.2B & 48.5 & 15.0 & 8.4 & 37.3 & 25.3 & 21.9 \\
\midrule
OCC-RAG-0.6B & 57.6 & 36.6 & 75.0 & \underline{79.9} & \underline{5.2} & 86.9 \\
OCC-RAG-1.7B & 60.9 & 38.2 & \textbf{81.0} & \textbf{81.4} & \textbf{5.0} & \underline{87.2} \\
\bottomrule
\end{tabular}}
\caption{Main evaluation results. ConFiQA metrics are averaged across QA, MR, and MC subsets. Parentheses (if available) represent the evaluation results with thinking mode on. We highlight the best result per column as \textbf{bold}, and second-best as \textbf{underline}.}
\label{tab:results}
\end{table}

%% file: appendices/trace_prompt.tex
\begin{figure}[htbp]
\begin{tcolorbox}[
    colback=gray!5!white,
    colframe=gray!75!black,
    title=Prompt,
    fonttitle=\bfseries,
    breakable,
    enhanced,
    fontupper=\ttfamily\footnotesize,
]
\ptok{query\_start}What is the name of the unit that the main character of the series featuring Karen Hayes is a member of?\ptok{query\_end}\\
\ptok{source\_start}\srcid{1} Karen Hayes is a fictional character on the television program ``24'' portrayed by actress Jayne Atkinson. She appeared as a recurring character in twelve episodes of the fifth season and a main cast member in eighteen episodes of the sixth season.\ptok{source\_end}\\
\ptok{source\_start}\srcid{2} Maxwell Greevey is a fictional character played by George Dzundza on NBC's long-running police procedural and legal drama television series ``Law \& Order''. Following Dzundza's departure from the cast at the end of the first season, Greevey was written off the series with his death in the second season premiere.\ptok{source\_end}\\
\ptok{source\_start}\srcid{3} Jack Bauer is a fictional character and the lead protagonist of the Fox television series ``24''. His character has worked in various capacities on the show, often as a member of the Counter Terrorist Unit (CTU) based in Los Angeles, and working with the FBI in Washington, D.C. during season~7.\ptok{source\_end}
\end{tcolorbox}

\begin{tcolorbox}[
    colback=gray!5!white,
    colframe=gray!75!black,
    title=Response,
    fonttitle=\bfseries,
    breakable,
    enhanced,
    fontupper=\ttfamily\footnotesize,
]
\rtok{query\_analysis\_start}\\
The question asks for the name of a specific unit. To find this, I need to: (1) identify the television series featuring the character Karen Hayes, (2) identify the main character of that series, and (3) determine the name of the unit that main character is a member of.\\
\rtok{query\_analysis\_end}\\[2pt]
\rtok{source\_analysis\_start}\\
\srcid{1}: states that Karen Hayes is a fictional character on the television program ``24''. This identifies the series.\\
\srcid{2}: discusses Maxwell Greevey and ``Law \& Order''; irrelevant to Karen Hayes.\\
\srcid{3}: states that Jack Bauer is the lead protagonist of ``24'' and that he often works as a member of the Counter Terrorist Unit (CTU).\\
\rtok{source\_analysis\_end}\\[2pt]
\rtok{reasoning\_start}\\
From \srcid{1}, the series featuring Karen Hayes is ``24''. From \srcid{3}, the main character of ``24'' is Jack Bauer, and Jack Bauer works as a member of the Counter Terrorist Unit (CTU). Therefore, the unit is the Counter Terrorist Unit (CTU).\\
\rtok{reasoning\_end}\\[2pt]
\rtok{status\_start} ANSWERABLE \rtok{status\_end}\\[2pt]
\rtok{answer\_start} Counter Terrorist Unit (CTU) \rtok{answer\_end}
\end{tcolorbox}
\caption{Example of the prompt/response format used at mid-training and at evaluation. The prompt wraps the question in {\color{ptokcolor}\texttt{<|query|>}} tokens and each context passage in {\color{ptokcolor}\texttt{<|source|>}} tokens prefixed by a numeric {\color{srccolor}\texttt{<|source\_id|>}}. The response is split into five sections delimited by {\color{rtokcolor}\texttt{<|*|>}} tokens; \emph{Reasoning} composes evidence from sources 1 and 3 into a three-hop chain (Karen Hayes $\rightarrow$ ``24'' $\rightarrow$ Jack Bauer $\rightarrow$ CTU), \emph{Status} carries the discrete refusal verdict, and \emph{Answer} carries the final span. The example is taken from the multi-hop, multi-context subset of the training corpus.}
\label{fig:trace-training-example}
\end{figure}

%% file: appendices/training_hyperparameters.tex
\begin{table}[H]
\centering
\small
\begin{tabular}{lcc}
\toprule
\textbf{Hyperparameter} & \textbf{OCC-RAG-0.6B} & \textbf{OCC-RAG-1.7B} \\
\midrule
Base model              & Qwen3-0.6B-Base   & Qwen3-1.7B-Base   \\
Optimizer               & AdamW             & AdamW             \\
Peak learning rate      & $1\times10^{-4}$  & $1\times10^{-4}$  \\
LR schedule             & cosine            & cosine            \\
Warmup ratio            & $0.03$            & $0.03$            \\
Weight decay            & $0.01$            & $0.01$            \\
Mixed precision         & bf16              & bf16              \\
Max sequence length     & $6{,}144$         & $6{,}144$         \\
Epochs                  & $1$               & $1$               \\
Per-device batch size   & $32$              & $16$              \\
Number of GPUs          & $8$               & $8$               \\
Global batch size       & $256$             & $128$             \\
Distributed strategy    & FSDP              & FSDP              \\
Memory-saving kernel    & Liger fused linear CE & Liger fused linear CE \\
Total training tokens   & $9\times10^9$     & $9\times10^9$               \\
Wall-clock training time (hours) & $17$              & $28$               \\
GPU type                & NVIDIA H100 (80~GB) & NVIDIA H100 (80~GB)               \\
\bottomrule
\end{tabular}
\caption{Mid-training hyperparameters for the released OCC-RAG checkpoints.}
\label{tab:training-hyperparameters}
\end{table}